\documentclass[runningheads]{llncs}

\usepackage[T1]{fontenc}

\usepackage{graphicx} 
\usepackage{algorithmic}
\usepackage{algorithm}
\usepackage{booktabs}
\usepackage{amsmath}
\usepackage{utfsym}
\usepackage{multirow}
\usepackage{enumitem}
\usepackage{amssymb}
\usepackage{colortbl}  
\usepackage{xcolor}
\usepackage{booktabs} 
\begin{document}
\title{DDSB: An Unsupervised and Training-free Method for Phase Detection in Echocardiography}

\titlerunning{DDSB for Phase Detection in Echocardiograph}

\author{
Zhenyu Bu\inst{2}\textsuperscript{*} \and Yang~Liu\inst{1}\textsuperscript{*†} \and
Jiayu Huo\inst{1} \and
Jingjing Peng\inst{1} \and
Kaini Wang\inst{2} \and
Guangquan Zhou\inst{2} \and
Rachel Sparks\inst{1} \and
Prokar Dasgupta\inst{1} \and
Alejandro Granados\inst{1} \and
Sebastien Ourselin\inst{1}
}
%
\authorrunning{Bu et. al.}

\institute{
King's College London, London, UK \and
Southeast University, China}

\maketitle  
\begingroup
\renewcommand\thefootnote{*}
\footnotetext{Equal contribution.}
\renewcommand\thefootnote{†}
\footnotetext{Corresponding author. Email: yang.9.liu@kcl.ac.uk}
\endgroup

\centerline{\tt\small \{yang.9.liu, firstname.lastname\}@kcl.ac.uk}
\centerline{\tt\small\{zybu, 230218244, guangquan.zhou\}@seu.edu.cn}
\begin{abstract}

Accurate identification of End-Diastolic (ED) and End-Systolic (ES) frames is key for cardiac function assessment through echocardiography. However, traditional methods face several limitations: they require extensive amounts of data, extensive annotations by medical experts, significant training resources, and often lack robustness. Addressing these challenges, we proposed an unsupervised and training-free method, our novel approach leverages unsupervised segmentation to enhance fault tolerance against segmentation inaccuracies. By identifying anchor points and analyzing directional deformation, we effectively reduce dependence on the accuracy of initial segmentation images and enhance fault tolerance, all while improving robustness. Tested on Echo-dynamic and CAMUS datasets, our method achieves comparable accuracy to learning-based models without their associated drawbacks. The code is available at~\url{https://github.com/MRUIL/DDSB}.

\end{abstract}
\section{Introduction}

\begin{figure}
\includegraphics[width=\textwidth]{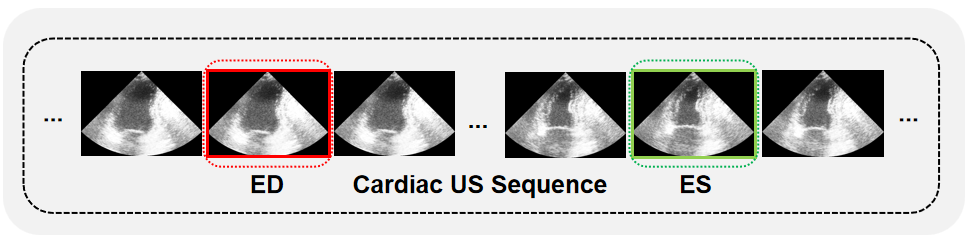}
\caption{Example of cardiac ultrasound sequence with ED in Red and ES in Green.} \label{fig1}
\label{fig:cardiac_cine}
\end{figure}

Cardiovascular diseases represent a major global health issue, underscoring the need for early detection~\cite{nabel2003cardiovascular}. As a cost-effective and real-time diagnostic technique, echocardiography plays a vital role in the diagnosis of these conditions~\cite{otto2013textbook}.
Accurately identifying the End-Diastolic (ED) and End-Systolic (ES) phases on echocardiograms, as shown in Fig.~\ref{fig:cardiac_cine}, is crucial for computing critical clinical metrics such as ejection fraction and global longitudinal strain. These metrics are critical in assessing cardiac functionality~\cite{halliday2021assessing}. Identifying ED and ES phases poses significant challenges, primarily due to the inherent variability in heart shapes, sizes and movement patterns among individuals. Moreover, the quality of echocardiographic images can be affected by factors such as imaging conditions and patient anatomy, further complicating the detection process.

%

%

Early techniques for identifying ED and ES frames in echocardiograms primarily relied on manual selection or simplistic automated criteria, often failing to accurately capture the heart's complex dynamics~\cite{mada2015define}. The innovative use of the QRS complex onset and the T wave's end as markers for ED and ES did not account for regional motion irregularities, proving to be impractical in emergency scenarios requiring rapid diagnosis~\cite{mada2015define}. Nadjia et. al.~\cite{kachenoura2007automatic} attempted to quantify the similarity between the ED and ES frames using the correlation coefficient, an approach that still required manual selection of the ED frame. Meanwhile, Barcaro et. al.~\cite{barcaro2008automatic}, along with Darvishi et. al.~\cite{darvishi2013measuring} and Abboud et. al.~\cite{abboud2015automatic}, explored automated segmentation techniques to delineate the left ventricle, identifying the ED and ES frames by the largest and smallest ventricular cross sections, respectively. However, these methods heavily depended on the accuracy of segmentation results and have a very poor tolerance for segmentation errors.

With the rapid advancement of deep learning technologies, methods for detecting ED and ES frames in echocardiograms have significantly evolved. These approaches generally fall into two main categories: classification-based~\cite{pu2021fetal} and regression-based~\cite{dezaki2018cardiac}. In classification models, ED, ES, and other frames are categorized into distinct labels, yet this often leads to class imbalance due to the singular occurrence of ED and ES frames within a series of intermediate frames. To counter this imbalance, some researchers have pivoted towards regression tasks, using interpolation to assign unique values to each frame, a strategy gaining popularity for its effectiveness in the field.
Further innovation has been seen in the application of Recurrent Neural Networks (RNNs)~\cite{elman1990finding}, traditionally used in natural language processing to understand character sequence correlations. Kong et. al.~\cite{kong2016recognizing} introduced TempReg-Net, integrating Convolutional Neural Networks (CNNs)~\cite{kim2014convolutional} with RNNs to pinpoint specific frames within MRI sequences, demonstrating a novel approach to leveraging deep learning for temporal and spatial feature extraction. Dezaki et. al.~\cite{dezaki2018cardiac} demonstrated how combining traditional CNN architectures with RNNs for temporal analysis, particularly employing DenseNet~\cite{huang2017densely} and GRU~\cite{chung2014empirical} models, could yield optimal results. This methodology, however, entails extensive trial and error to identify the most effective CNN and RNN combinations.
Wang et. al.~\cite{wang2021simultaneous} proposed a dual-branch feature extraction model, defining the task of identifying ED/ES frames as a curve regression problem, thus moving away from direct index regression. Meanwhile, Li et. al.~\cite{li2023semi} explored a semi-supervised approach for ED/ES detection, requiring only a portion of labeled data, thereby reducing the dependency on extensively annotated datasets. Singh et. al.~\cite{singh2023preprocessing} advanced the methodology by combining CNN with Bidirectional Long Short-Term Memory (BLSTM) networks~\cite{zhang2015bidirectional}, offering an enhanced solution over previous models.
These deep learning strategies require extensive annotated datasets. Acquiring such data and securing annotations from medical professionals present substantial challenges due to the lack of resources and the intensive workload required.

To overcome the above drawbacks, we proposed an unsupervised and training-free method, called \textbf{DDSB} (Directional Distance to Segmentation Boundary), for phase detection.
Our contributions are three-fold:

\begin{itemize}[label=$\blacktriangleright$]
    \item We have proposed a novel unsupervised and training-free approach for phase detection, which can recognize the ED/ES phase in the cardiac cine without the need for annotated datasets for training and avoid the GPU resource wastage caused by the training process.
    
    \item We employed a distance-based strategy to formulate proposed modifications from diverse perspectives, utilizing initial segmentation outcomes and serving as the frame representation. This approach is designed to fight the inherent limitations of coarse segmentation results, thereby strengthening the robustness of our model.

    \item We demonstrate the effectiveness and advantages of our method on two public datasets (Echo-dynamic~\cite{ouyang2020video} and CAMUS~\cite{leclerc2019deep}), achieving comparable performance compared to other deep learning based approaches.
    
\end{itemize}

\section{Method}

\begin{figure}[t]
\centering
\includegraphics[width=0.8\textwidth]{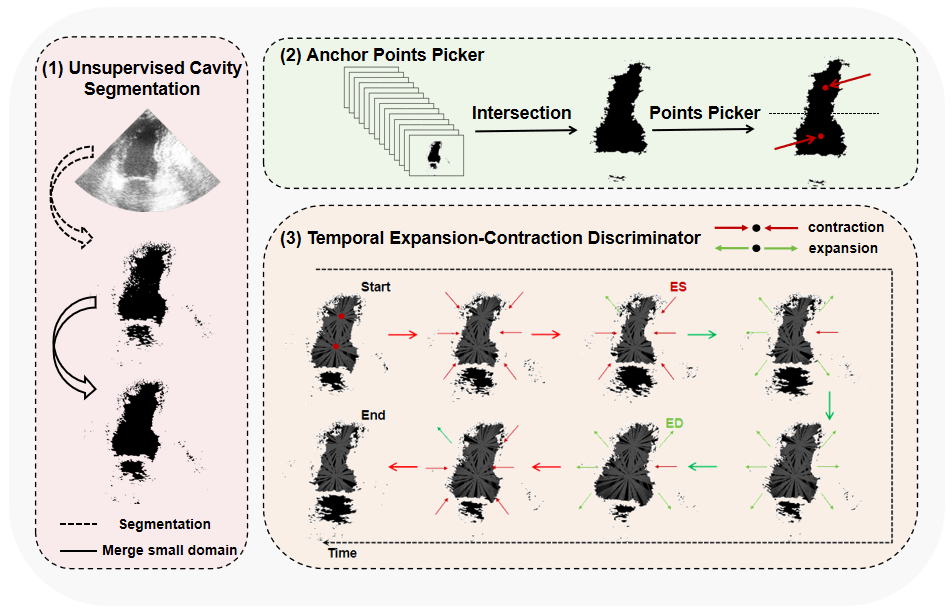}
\caption{ %
The overview of our DDSB for unsupervised and training-free ED/ES detection.%
} \label{fig:framework}
\vspace{-1mm}
\end{figure}

In this work, we delve into the challenge of unsupervised detection of echocardiography phases, with a keen focus on identifying the End-Diastolic (ED) and End-Systolic (ES) stages. Our objective is to devise an algorithm \textit{f} that can reliably approximate the moments of ED and ES, denoted as $(t_{ed}, t_{es})$, from a given sequence of \(T\)-frame echocardiographic video, $X_T = { \boldsymbol{x}_j }_{j = 1}^{T}$. Here, $\boldsymbol{x}_j$ signifies the $j$-th frame within the video stream. The proposed methodology, referred to as DDSB (Directional Distance to Segmentation Boundary), encapsulates three integral components, as shown in Fig.~\ref{fig:framework}: an unsupervised segmentation optimizer, an anchor points picker, and a temporal expansion-contraction discriminator. DDSB stands out as an innovative, unsupervised, and training-free approach in the realm of cardiac phase detection.

\subsection{Unsupervised cavity segmentation}


For the effective identification of ED and ES frames in echocardiography videos, the primary step involves the precise segmentation of interested regions, specifically, the heart chambers. We achieve this using an unsupervised adaptive threshold segmentation algorithm~\footnote{`cv2.adaptiveThreshold()' function as the OpenCV package, which adjusts thresholds based on local intensity variations to accurately segment cardiac structures}. The result of this process is an initial segmentation sequence, $\mathcal{S}_i$, where the ventricular cavities are marked as `0' (indicating the absence of cardiac tissue), and the myocardial along with other areas are marked as `1' (indicating their presence).

To address the challenge posed by noise within the ventricular cavities, which can significantly disrupt further processing, we apply a threshold, $s$, to filter based on the area of the connected non-cavity regions. The process for obtaining the filtered segmentation, $\mathcal{S}_{f}$, is succinctly formulated as follows:
\[
\mathcal{S}_{f}(p) = \mathcal{S}_i(p) \cdot \mathbf{1}_{\{\mathcal{A}(p) \geq s\}},
\]
where $p = (x,y)$ denotes the pixel coordinates, $\mathcal{A}(p)$ quantifies the area of the connected non-cavity domain at pixel $p$, and 
$\mathbf{1}_{\{\cdot\}}$ equals 1 if `$\cdot$' is true, otherwise it is 0. This strategy ensures that $\mathcal{S}_{f}$ achieves a refined segmentation by proficiently minimizing noise impacts.

\subsection{Anchor points picker}
For effective analysis of cardiac chamber dynamics, especially to distinguish the dilation and contraction phases, identifying anchor point(s) $P_a$ within the heart chamber is crucial. These points act as references for observing boundary movements in relation to $P_a$. Initially, we select these points based on their persistent presence within the cardiac cavity across the video sequence. Hence, the potential positions, $M_a$, are identified as:
\[
M_a = \left\{ p \mid \sum_{i=1}^{T} \mathcal{S}_f(i, p) \leq \text{Percentile}\left( \sum_{i=1}^{T} \mathcal{S}_f(i, \cdot), 1\% \right) \right\},
\]
where $\text{Percentile}(\cdot, 1\%)$ calculates the 1\% percentile over the set of summed values. To ensure that $P_a$ is located within the cardiac cavity and not in other cavities, we further refine our selection. Among the four largest connected domains by area, we opt for the one nearest to the image's top center $p_{top} = (0,\frac{W}{2})$, where $W$ is the frame width, as the final anchor point candidate. This decision takes advantage of the proximity of the center of the domain to the top center of the image, ensuring an optimal representation of the cardiac cavity:
\[
P_a = \underset{\mathcal{C} \subseteq \mathcal{L}_4(M_a)}{\text{argmin}} \, d(\text{Center}(\mathcal{C}), \, p_{top}),
\]
with $\mathcal{L}_4(M_a)$ representing the four largest connected domains within $M_a$, and $d(\cdot,\cdot)$ denoting the distance between two points.

The ventricular center point serves as an optimal reference due to its stability and unobstructed perspective on various boundary locations, minimizing the likelihood of occlusion. To fully capture the deformation, the incorporation of additional anchor points is beneficial. Given the predominantly longitudinal configuration of the heart, we propose a strategy to partition $P_a$ into $t_a$ vertical segments. From each segment, we select a central point as an anchor, thereby acquiring $t_a$ anchor points. We divide $P_a$ into $t_a$ regions vertically, denoted by $\{R_1, R_2, \ldots, R_{t_a}\}$. For each region $R_i$, we determine the central anchor point $C_i$ as:
\[
C_i = \text{Center}(R_i), \quad \text{for } i = 1, 2, \ldots, t_a,
\]
where $\text{Center}(R_i)$ computes the centroid of region $R_i$. Consequently, the set of anchor points $\{C_1, C_2, \ldots, C_{t_a}\}$ provides a detailed representation of cardiac deformation, enhancing the analysis of ventricular dynamics.

\subsection{Temporal expansion-contraction discriminator}

To elucidate heart deformation, we introduce the \textit{change description element}, \(\delta(\theta, C_i, j)\), to quantify boundary distance changes between consecutive frames \((x_j, x_{j+1})\) along a specific direction \(\theta\), using the reference point \(C_i\). Positive values of \(\delta\) signify deformation, indicating expansion or contraction of the heart. Given the potential challenges associated with the fidelity of unsupervised segmentation, we enhance the method's robustness by analyzing \(k\) directions, equally spaced, from each anchor point. Consequently, for each pair of adjacent frames, we compile:
\[
L_i = \left\{\delta\left(\frac{2\pi k_0}{k}, C_i, j\right) \mid k_0 \in \{1, \ldots, k\}, C_i \in \{C_1, \ldots, C_{t_a}\}\right\}.
\]
Detecting a significant change in \(\delta\) between two frames that exceeds a predefined threshold \(\alpha\) is considered an anomaly. The expansion rate between frames is calculated by balancing the sums of positive and negative \(\delta\) values, normalized by the total count of valid \(\delta\):
\[
E_j = \frac{\sum_{\alpha > \delta(\cdot,\cdot,j) > 0 }   -\sum_{ -\alpha<\delta(\cdot,\cdot,j) < 0}}{\sum_{|\delta(\cdot,\cdot,j)| < \alpha} + 1e-6}.
\]
A negative \(E_j\) implies contraction, offering an immediate assessment of deformation dynamics between the frames.

To accurately capture the heart's relative size at a specific frame \(j\), we define \(A_j = \sum_{i=1}^j E_i\), where \(E_i\) represents the expansion rate between consecutive frames. This summation reflects the heart's cyclical pattern of expansion and contraction. Identifying the ED/ES phases is achieved by pinpointing the indices \((i_0, j_0)\) that maximize the absolute difference in cumulative expansion rates, as given by:
\[
(i_0, j_0) = \underset{i<j}{\mathrm{argmax}} \left|2A_{i} - 2A_{j} + A_T\right|,
\]
where \(A_T\) is the total cumulative expansion rate at the final frame. The criterion \(2A_{i_0} - 2A_{j_0} + A_T > 0\) indicates an initial phase of contraction followed by expansion, denoting \((i_0, j_0)\) as \((t_{ed}, t_{es})\). Conversely, a negative value suggests an initial expansion followed by contraction, leading to \((t_{ed}, t_{es}) = (j_0, i_0)\). This methodology adeptly delineates the ED and ES phases by examining the heart deformation pattern across the sequence.

\section{Dataset and Metrics}

\subsubsection{Dataset}

In our research, we used the Echo-dynamic and CAMUS datasets for validation. However, we needed to adjust them for our experiments. For CAMUS, where ED and ES phases are at the start and end of the cycle, we mixed up the sequence to prevent models from just focusing on these endpoints. We did this by randomly selecting two points in the cycle, flipping the frames between them to form a varied sequence.
In the Echo-dynamic dataset, where only one ED/ES pair is labeled, we removed sequences with too short intervals between these phases for consistency. Then, we chose sequences that included the labeled ED/ES pair and added variability by cutting the sequence at two random points.
Regarding sequence length, the original CAMUS dataset had 500 samples, each 10 to 32 frames long. After our modifications, the average sequence length increased to about 36 frames, making the data more suitable for our analysis.

\subsubsection{Metrics}

In our evaluation, we measure accuracy using the mean absolute error (MAE) in frames, denoted by \(\mu\). This is calculated as the average absolute difference between the predicted frame index (\(\tilde{t}\)) and the true frame index (\(t\)) across all \(N\) samples in the dataset, specifically for the ED or ES frames.

\section{Results and Discussion}
\subsection{Comparisons with State-of-the-art Methods}

\begin{table}[t]
\centering
\caption{Comparative analysis of experimental results: A perspective on the necessity of training and the application of supervised learning approaches}
\label{tab:comparison_with_sota}
\resizebox{\linewidth}{!}{
\begin{tabular}{ccccccc}
\toprule
Datasets (Training Set Num.)   &Dataset Scale   & Methods & Supervised & Trained  & $\mu_{ED} \downarrow$ &$\mu_{ES} \downarrow$ \\ 
\midrule
\multirow{7}{*}{CAMUS (450)}  &\multirow{4}{*}{100\%}     & Kong et. al.~\cite{kong2016recognizing} &  \usym{1F5F8}    &  \usym{1F5F8}  &1.59&2.31\\
&             & Dezaki et. al. \cite{dezaki2018cardiac} &  \usym{1F5F8}    & \usym{1F5F8}  &1.44&1.99\\
&             &  Singh et. al.\cite{singh2023preprocessing}&  \usym{1F5F8}   &  \usym{1F5F8}  & 1.77 & 2.59\\
&            & Li et. al.\cite{li2023semi}&  \(Semi\)    & \usym{1F5F8}  &2.23&2.73\\ 
             \cline{2-7}
& 50\%            & Dezaki et. al. \cite{dezaki2018cardiac} &  \usym{1F5F8}    & \usym{1F5F8}  &2.97&3.48\\
\cline{2-7}
  & \multirow{2}{*}{0\%}           &  Size-based & \usym{2717}   &  \usym{2717}  & 6.09 & 3.68\\
 &            & DDSB (Ours)&  \usym{2717}     &  \usym{2717}  &2.27&1.29\\
\midrule
\midrule
\multirow{7}{*}{Echo-dynamic (5970)} &\multirow{4}{*}{100\%} &  Kong et. al.\cite{kong2016recognizing}&  \usym{1F5F8}   &  \usym{1F5F8}  &1.35&2.76  \\
  &           &  Dezaki et. al. \cite{dezaki2018cardiac}&  \usym{1F5F8}   &  \usym{1F5F8}   &1.32&2.04        \\
   &          &  Singh et. al.\cite{singh2023preprocessing}&  \usym{1F5F8}   &  \usym{1F5F8}  &1.98&2.56\\
    &        &  Li et. al.\cite{li2023semi}&  \(Semi\)    &  \usym{1F5F8}  &2.10&1.70\\
              \cline{2-7}
    & 50\%            & Dezaki et. al. \cite{dezaki2018cardiac} &  \usym{1F5F8}    & \usym{1F5F8}  &4.12&3.79\\
    \cline{2-7}
        & \multirow{2}{*}{0\%}       &  Size-based &  \usym{2717}   &  \usym{2717}  &12.54 &9.24\\
    
          &   &   DDSB (Ours) &  \usym{2717}    &  \usym{2717}        &3.84&4.62\\
\bottomrule
\end{tabular}
}
\end{table}

\begin{figure}[t]
\centering
\includegraphics[width=0.9\textwidth]{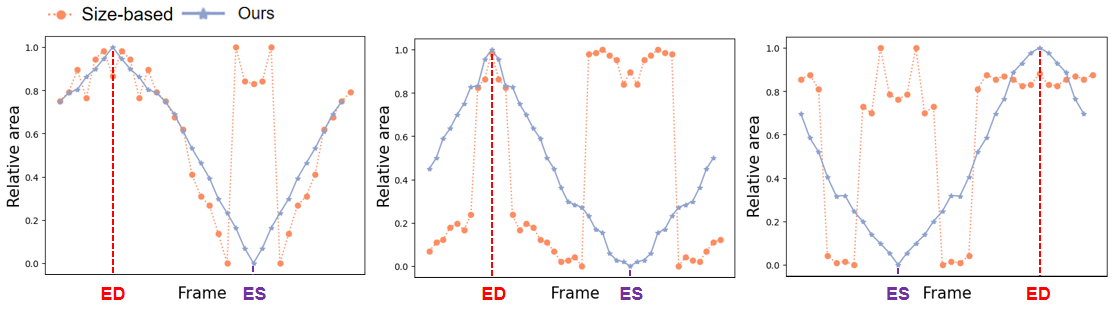}
\caption{Comparison of the size-based method with our DDSB.} \label{fig:visualization}
\vspace{-1mm}
\end{figure}

\subsubsection{In-dataset Evaluation.} We compared methods for detecting ED and ES in echocardiograms on the CAMUS and Echo-dynamic datasets. We rebuilt each deep learning model from the literature for fair comparison and set a baseline using a size-based approach, the size of the connected area with our anchor point, as shown in Tab~\ref{tab:comparison_with_sota}. On CAMUS, our unsupervised method not only competes well but also beats the top supervised model by 0.7 in \(\mu_{ES}\). Even with half the training data, it still surpasses Dezaki et al.'s model, reducing ED and ES errors by 0.7 and 2.19 frames, respectively. Compared to the size-based baseline, our method significantly cuts down errors by 3.82 frames for ED and 2.39 for ES. On the Echo-dynamic dataset, our method is slightly less effective than deep learning models trained with 5970 samples. However, with a smaller dataset of 2985 samples, our performance matches the best models. This highlights our method's value in scenarios with limited labeled data.

We compared our method to the size-based one through visuals in Fig.~\ref{fig:visualization}. Our method deals better with errors in the mask and shows fewer, smoother changes, proving its reliability. Overall, it is more precise and stable than the size-based method. Also, unlike deep learning methods that only offer a final result, our method can show the heart's changes dynamically, making it more flexible for different uses.
\subsubsection{Cross-dataset Evaluation.}
We tested our method's generalization against Dezaki et al.'s by training on one dataset and testing on another. Our method showed better accuracy in this cross-dataset setting, while it was slightly less effective in in-dataset evaluations, highlighting our strong generalization compared to deep learning-based methods in this task.

\begin{table}[t]
\renewcommand\arraystretch{1.2}
\centering
\caption{Cross-dataset evaluation results. We evaluate performance by testing a model trained on one dataset on a different dataset.
}
\label{tab:crossdatasetvalidation}
\resizebox{0.6\linewidth}{!}{
\begin{tabular}{ccc|cc}
\toprule
& \multicolumn{2}{c|}{CAMUS $\rightarrow$ Echo-dynamic } & \multicolumn{2}{c}{Echo-dynamic $\rightarrow$ CAMUS} \\
\cline{2-5}
Metrics & $\mu_{ED}$ & $\mu_{ES}$ & $\mu_{ED}$ & $\mu_{ES}$ \\
\midrule
Dezaki et. al.~\cite{dezaki2018cardiac} & 4.24 & \textbf{3.99} & 3.13 & 4.21 \\
DDSB (Ours) &\textbf{3.84} &4.62 &\textbf{2.27} &\textbf{1.29} \\
\bottomrule
\end{tabular}
}
\vspace{-2mm}
\end{table}

\subsection{Ablation Study}

We conducted ablation studies on the CAMUS dataset to understand how certain hyper-parameters affect our method.

\subsubsection{Effects of $k$ directions.}
We calculated $k$ directional distances from each anchor point to the segmentation boundary. Our findings, shown in Tab.~\ref{tab:sample-table}, indicate that our method's performance is stable across different $k$ values, highlighting its robustness.

\subsubsection{Effects of change threshold $\alpha$.}
We defined any distance change between adjacent frames exceeding $\alpha$ as invalid. This approach led to an improvement of approximately 0.5 for $\mu_{ED}$ and 0.17 for $\mu_{ES}$, as documented in Tab.~\ref{tab:alpha} Our method's effectiveness is not heavily dependent on the exact value of $\alpha$, showcasing its flexibility.

\begin{minipage}{\textwidth}
\begin{minipage}[t]{0.45\textwidth}
\makeatletter\def\@captype{table}
\caption{Effects of $k$ directions.}
\centering
\setlength{\tabcolsep}{6.6pt}
\renewcommand{\arraystretch}{0.9}
\resizebox{0.8\linewidth}{!}{
\begin{tabular}{c|c|c}
    \toprule
    $k$     & $\mu_{ED} \downarrow$     & $\mu_{ES} \downarrow$ \\
    \midrule
    72     &2.36&1.46      \\
    180     &2.63&1.39      \\
    360     &2.75&1.62      \\
    \bottomrule
\end{tabular}
}
\label{tab:sample-table}
\end{minipage}
\hspace{2mm}
\begin{minipage}[t]{0.45\textwidth}

\makeatletter\def\@captype{table}
\caption{Effects of change threshold $\alpha$.}
\centering
\setlength{\tabcolsep}{6.6pt}
\renewcommand{\arraystretch}{0.9}
\resizebox{0.6\linewidth}{!}{
\begin{tabular}{c|c|c}
    \toprule
    $\alpha$     & $\mu_{ED} \downarrow$     & $\mu_{ES} \downarrow$ \\
    \midrule
    None    &2.77&1.46      \\
    \midrule
    5     &\textbf{2.27}&1.29     \\
    10     &2.36&1.46      \\ 
    15     &2.57&\textbf{1.28}      \\ 
    \bottomrule
\end{tabular}
}
\label{tab:alpha}
\end{minipage}
\end{minipage}

\section{Conclusion}
We introduced DDSB, an innovative unsupervised and training-free method for phase detection. Our approach yields results that are on par with the latest supervised deep learning methods but without the need for training resources, such as computational power, data labeling, or extensive datasets. This advantage becomes particularly clear when data availability is limited. Moreover, DDSB allows for dynamic visualization of results, unlike deep learning methods that offer a single result, broadening its application. DDSB does not always outperform supervised learning with certain datasets. Yet, our framework has the potential to integrate with deep learning techniques, offering new perspectives for improving them, which is also in our plan.

\bibliographystyle{splncs04}
\bibliography{mybibliography}

\end{document}